# Par4Sim – Adaptive Paraphrasing for Text Simplification

**Seid Muhie Yimam      Chris Biemann**

Language Technology Group
Department of Informatics, MIN Faculty
Universität Hamburg, Germany
{yimam,biemann}@informatik.uni-hamburg.de

## Abstract

Learning from a real-world data stream and continuously updating the model without explicit supervision is a new challenge for NLP applications with machine learning components. In this work, we have developed an adaptive learning system for text simplification, which improves the underlying learning-to-rank model from usage data, i.e. how users have employed the system for the task of simplification. Our experimental result shows that, over a period of time, the performance of the embedded paraphrase ranking model increases steadily improving from a score of 62.88% up to 75.70% based on the NDCG@10 evaluation metrics. To our knowledge, this is the first study where an NLP component is adaptively improved through usage.

## Title and Abstract in Amharic

Par4Sim -- የሚጣጣም መልሶ መጻፍ፣ ፅሁፍን ለማቅለል

ከገዛዱ ዓለም የመረጃ ፍሰት መማርና ቀጥተኛ ድጋፍ ሳያስፈልገው ሞዴሉን በተከታታይ ለማሻሻል ማሽን ለርኒንግ (በራሱ መማር የሚችል የኮምፒተር መርሃ ግብር) በሚያካትቱ የተፈጥሯዊ የቋንቋ ቴክኖሎጂ (ተ.ቋ.ቴ - NLP) ጥናት አዲስ ፈተና ሆኗል። በዚህ ምርምር፣ ተጠቃሚዎች እንድን መተግበሪያ በመጠቀም ላይ ሳሉ ማለትም በተጠቃሚዎች የተለዩ ድርጊቶች መሰረት ውህቦችን ከበስተጀርባ ሆኖ በመሰብሰብ የማሽን ለርኒንጉን ሞዴል በማዘመን እየተጣጣመ መማር የሚችል የጽሁፍ አቅላይ ስርዓት ሰርተናል። የሲስተማችንን ውጤታማነት ለማረጋገጥ፣ እንድን ፅሁፍ ለተለያዩ ተጠቃሚዎች በቀላሉ እንዲገባቸው ለማድረግ ተጣጣሚ ትርጉሜዎችን እንደ ምሳሌ ወስደናል። በመሆኑም፣ አንድ ተጠቃሚ እንድን ፅሁፍ መረዳት ሲያቅተው፣ የኛን ተንታኝ መተግበሪያ ጠቅሞ አጋራጭ ቃል ወይንም ሃረግ በመተካት ፅሁፉን የበለጠ ቀላል ማድረግ ይችላል።

የሙከራ ውጤታችን እንደሚያሳየው ከሆነ፣ በተከታታይ ጊዜ ውስጥ እንዲማር/እንዲላመድ በሲስተሙ ውስጥ የተሸጎጠው ደረጃ መዳቢ ሞዴል፣ በNDCG@10 የግምገማ ልኬቶች መሰረት፣ ሳያቋርጥ በመጨመር ከ62.88% እስከ 75.70% ድረስ መሻሻል አሳይቷል። እኛ እስከምናውቀው ድረስ፣ ይህ ጥናት፣ የተፈጥሯዊ የቋንቋ ቴክኖሎጂ (ተ.ቋ.ቴ - NLP) ትግባሪዎች በአግልግሎት አጠቃቀም ጊዜ ውህቦችን በመሰብሰብ የማሽን ለርኒንግ ስልት ቀምር ሞዴል አገልግሎት በሂደት ውስጥ እያለ እንዲሻሻል/እንዲላመድ የሚያደርግ የመጀመሪያው ጥናት ነው።

## 1 Introduction

Traditionally, training data for machine learning-based applications are collected amass, using a specific annotation tool. There are a number of issues regarding this approach of data collection. Most importantly, if the behavior of the target application changes over time, this makes the training data outdated and obsolete, an issue known as *concept drift* (Žliobaitė et al., 2016).

In this regard, we opt to design an approach where data can be collected interactively, iteratively and continuously using an adaptive learning model in a live and a real-world application. By adaptive, we mean that the learning model gets signal from the usage data over a period of time and it automatically adapts to the need of the application. An adaptive learning model has a spectrum of advantages. First of all, the model gets updated continuously. The model also provides suggestions through usage and the user can correct the suggestions, which in turn improves the model's performance. On top of this, there is no need to collect a large set of training examples a priori, which might be difficult and expensive. We also believe that instead of collecting training data in advance, it is a more natural way to get the training examples from usage data by embedding the adaptive model in the application.

In this premise, we choose the advanced NLP task of text simplification in order to experiment with an adaptive learning approach. Text simplification aims at reducing the complexity (syntactic and lexical) of a text for a given target reader. According to the survey by Shardlow (2014), different approaches such as lexical simplification, explanation generation, and machine translation are used for text simplification. As far as our knowledge is concerned, there is no work regarding the use of an adaptive model, more specifically *adaptive paraphrasing based on usage data* for the task of text simplification.

The main technical challenge is the integration of an adaptive paraphrasing model into a text simplification writing aid tool in order to be able to attain whether it is possible to gain NLP component quality through adaptive learning on usage data. The writing aid tool for text simplification using an adaptive paraphrasing model (Par4Sim) consists of several components. The first component in the pipeline determines the complex or difficult words or phrases (CPs), which is based on the work of Yimam et al. (2017). Once the CPs are identified, the second component produces candidate suggestions from different paraphrase resources. Candidates that do not fit the context in the sentence are filtered or excluded based on a language model score. Finally, an adaptive ranking model reorders candidates based on their simplicity and provides the ranked candidates to the user within an interactive writing aid tool.

This work is aiming to answer the following three research questions:
*RQ1*: How can an adaptive paraphrase ranking model be integrated into a text simplification writing aid tool?
*RQ2*: How can an adaptive paraphrase ranking model be evaluated?
*RQ3*: Can we demonstrate the adaptivity of the approach?

This paper is organized as follows. In Section 2 we briefly review state-of-the-art works in adaptive learning and text simplification. Section 3 outlines the design procedures we have followed in the development of the Par4Sim tool for the integration of adaptive paraphrasing in a text simplification application in detail. In Section 4, a brief description of the data collection and statistics of the collected data is presented. Section 5 describes the learning-to-rank machine learning algorithm employed to build an adaptive ranking model, including the definition of employed feature representation and the evaluation metrics. The experimental results obtained and the contributions of our research work are presented in Section 6 and Section 7. Finally, Section 8 presents the conclusions of our work and indicates future directions on the integration of adaptive approaches for NLP applications.

## 2 Related Work

Supervised machine learning approaches commonly rely on the existence of a fixed training set, which is collected in advance.

The survey by Parisi et al. (2018) indicates that *continual lifelong learning*, the ability to learn continuously by acquiring new knowledge is one of the challenges of modern computational models. The survey further explains that one of the main problems of continual learning is that training model with new information interferes with previously learned knowledge. The work of To et al. (2009) indicates that the integration of user feedback in an adaptive machine learning approach helps to improve the system's performance. It further shows that the approach helps to reduce the manual annotation efforts. Žliobaitė et al. (2016) revealed that supervised machine learning approaches stationed with static datasets face problems during deployment in a real-world application due to concept drift (Tsymbal, 2004) as applications start generating data streams continually. Applications with such properties include spam filtering and intrusion detection.

Stream-based learning and online learning (Bottou, 1998) are alternative setups to a batch-mode adap-tive learning system. For example, the work by Levenberg et al. (2010) shows that the deployment of stream-based learning for statistical machine translation improves the performance of their system when new sentence pairs are incorporated from a stream. The work by Wang et al. (2015) presents SOLAR, a framework of scalable online learning algorithms for ranking, to tackle the poor scalability problem of batch and offline learning models. The work by Yimam et al. (2016a) uses MIRA (Margin Infused Relaxed Algorithm) (Crammer and Singer, 2003), a perceptron-based online learning algorithm to generate suggestions in an iterative and interactive annotation setup for a biomedical entity annotation task.

Most text simplification approaches employ basic machine translation models from parallel corpora (Xu et al., 2016; Štajner et al., 2017) and using simple Wikipedia for English (Coster and Kauchak, 2011). Simple PPDB (Pavlick and Callison-Burch, 2016) is such a resource that is built automatically, using machine translation techniques on a large number of parallel corpora.

The work by Lasecki et al. (2015) shows that using crowdsourcing for text simplification is a valid approach. We also conducted our adaptive text simplification experiment on the Amazon Mechanical Turk crowdsourcing platform using a specialized text simplification tool.

## 3 Design of the Par4Sim System

Par4Sim is a text simplification tool based on an adaptive paraphrasing paradigm. Unlike the traditional text simplification approaches, Par4Sim mimics a normal text editor (writing aid tool). The tool embeds basic text simplification functionalities such as providing suggestions for complex phrases, allowing editing of the texts in place and so on. Our setup is that authors wishing to simplify text can use the tool, without a priori machine learning model, but the system learns from the user interactions in an iterative and interactive manner. In order to run a distributed and web-scale simplification experiment, we have integrated the tool into an existing crowdsourcing platform.

Hence, our targeted users for the Par4Sim experiments are workers from the Amazon Mechanical Turk (MTurk) crowdsourcing platform. As a large number of workers are available in the MTurk platform, we paid special attention in the development of the tool regarding response time, accessibility, and reliability. The tool comprises a front-end component to edit texts and a back-end component, which exposes most of the requests using REST API services. Figure 1 shows the main components of the tool. The detailed instructions for the MTurk task are displayed in Figure 2.

While it is impossible to host complex systems inside the MTurk infrastructure, MTurk supports external human intelligence tasks (HITs) where workers can easily access our system through the MTurk interface. Par4Sim's user interface is embedded into the MTurk web page, but every activity is handled by our own server. This design gives us full control on the collection of the dataset, for training new ranker models, and for updating the model iteratively and seamlessly while we still use the MTurk infrastructure to recruit workers and pay rewards for a web-scale experiment.

As it can be seen in Figure 1, complex words or phrases (CPs) are automatically highlighted (yellow background color and underlined in cyan color). The highlighting of the CPs is based on the work of Yimam et al. (2017). Furthermore, the users can highlight their own CPs and our system will provide ranked candidates. Details about the generation of candidates and the ranking model are provided in Section 4. Besides the highlighting of CPs and providing ranked candidates, the system further provides the following functionalities.

- **Reload text**: If the worker wants to get the original text with the CPs re-highlighted, she can reload the content using the *Reload* button, subject to confirmation.

- **Undo and redo**: At any particular time, workers can undo or redo the changes they have made using the *Undo* and *Redo* buttons.

- **Highlight difficult words**: It is also possible to request our system to automatically highlight difficult words. This functionality is particularly important if the worker has changed the original text but she is not sure if the amended text is in fact simpler. Once the system highlights some words or phrases, she can still check if the suggestions provided by the system could still simplify the text further.

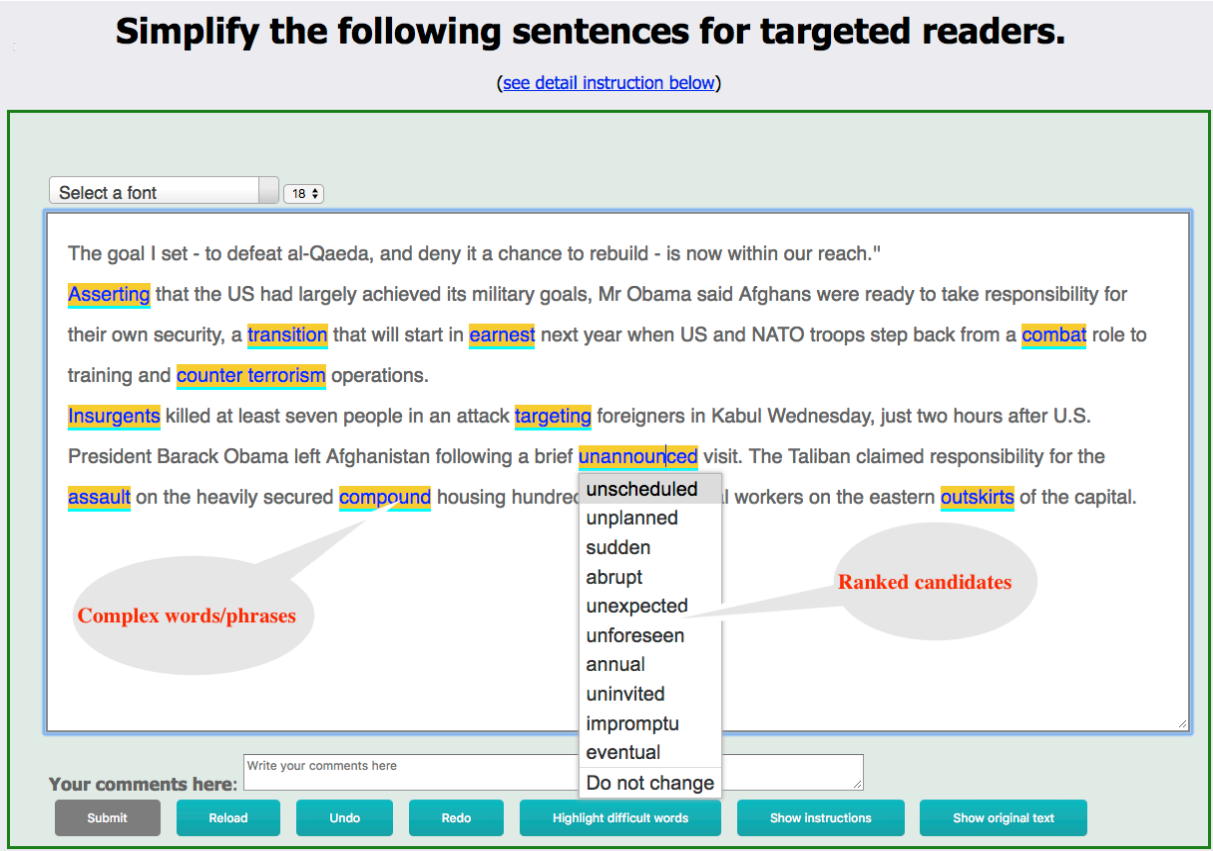

Figure 1: The Par4Sim UI as it is displayed inside the MTurk webpage with the instruction *"Simplify the following sentences for targeted readers"*. The *targeted readers* are explained in the detailed instruction as *children*, *language learner*, and *people with reading impairments*.

- **Show instruction / Original texts**: The instructions (see Figure 2) are visible at the beginning of each task. Once workers have accepted the task, the instructions will be hidden automatically so that the workers can focus on the simplification task in a clean window. During simplifying the text, the worker can display the instructions below the text editor if she wants to refer to the instructions. Workers can also compare the simplified version of the text with the original text.

- **Show animation**: Crowdworkers prefer a very short task description or the task should be easy enough to be understood by most workers. Since the simplification task is difficult, we decided to include an animated video showing important steps in the text simplification process. The video animation starts as soon as the task description is displayed and is hidden once the task is started. The worker can refer to the animation while completing the task.

With these functionalities, we provided a text simplification aid that comes very close to how a simplification application would look like. The goal was to provide a realistic environment in order to test the adaptive approach within a user-centric scenario. While we embedded the application into MTurk in order to attract paid users, it is straightforward to provide this web-based application online or locally.

## 4 Task Description and Dataset

In this paper, we address text simplification, which is the task to simplify a given text that is assumed to be difficult to understand for particular readers such as language learners, children or people with reading impairments. A text simplification pipeline usually starts with the complex word or phrase identification.

Figure 2: The detailed instructions of Par4Sim, which are displayed at the beginning of the task and hidden afterwards. The worker can display the instruction when required.

For this experiment, we have used parts of the dataset from Yimam et al. (2017), which already contains manually identified complex phrases (CP). In this dataset, the complex phrases are manually identified by 10 native and 10 non-native English speakers. The dataset has been already used for the complex word identification (CWI) shared task 2018[1]. Please refer Yimam et al. (2017) for the details of the dataset. We purposely used the manually identified and verified CPs because 1) we do not want to mix the identification and the simplification tasks, and 2) we want to test the adaptive learning approach in a controlled setting.

We have generated candidate suggestions from different paraphrase resources. The following resources are used to generate candidate suggestions:

- **Lexical and Distributional resources**: We use WordNet (Miller, 1995) and distributional thesaurus (Biemann et al., 2013) to produce candidate results for CPs. We apply lemmatization to reduce the CPs into their base forms and retrieve the top 10 synonyms respectively the top 10 similar words from the lexical resources.

- **PPDB 2.0 and Simple PPDB**: PPDB (Pavlick et al., 2015) is the largest paraphrase resource to date. The recently released simple PPDB (Pavlick and Callison-Burch, 2016) is a particularly relevant resource for the task of text simplification. For each CPs (source entry in PPDB), we retrieve the top 10 candidates (target entry in PPDB).

- **Phrase2Vec**: We have trained a Phrase2Vec model (Mikolov et al., 2013) using English Wikipedia and the AQUAINT corpus of English news text (Graff, 2002). Mikolov et al. (2013) pointed out that it is possible to extend the word-based model to a phrase-based model using a data-driven approach where each phrase or multi-word expressions are considered as individual tokens during the training process. We have used a total of 79,349 multiword expression and phrase resources, which are obtained from the work of Yimam et al. (2016b). We trained the Phrase2Vec embeddings with 200 dimensions using skip-gram training and a window size of 5. We retrieve the top 10 similar words to the CPs as candidates.

---
[1] https://sites.google.com/view/cwisharedtask2018/ (Yimam et al., 2018)

Obviously, the number of candidate suggestions obtained from these different resources is enormous and we should limit the size before providing to the ranker model. The candidates are ordered by a language model score. We trained a tri-gram language-model (Pauls and Klein, 2011) using the Wikipedia articles. The number of candidates is limited to 10; these are re-ranked using the learning-to-rank model.

For each HIT, we provide between 5 and 10 sentences for simplification. A HIT is then assigned to 10 workers as we need a graded relevance to train the learning-to-rank model (see Section 5). In the experiment, we make sure that a HIT is submitted only in one iteration and most importantly, during the evaluation of the ranking model performance, we make sure that the training data from previous iterations should not contain HITs from the current iteration.

In this experiment, a total of 18,036 training instances have been collected. Figure 3 shows how the training instances collected from the usage data looks like. The number at the end of the simplified sentence shows the number of workers provided the same simplified sentence.

**Complex Sentence:** *Hajar said his cousin was not affiliated with any terrorist group.*
**Simplified Sentence 1:** *Hajar said his cousin was not associated with any terrorist group.* → *6*
**Simplified Sentence 2:** *Hajar said his cousin was not merged with any terrorist group.* → *2*
**Simplified Sentence 3:** *Hajar said his cousin was not aligned with any terrorist group.* → *1*
**Simplified Sentence 4:** *Hajar said his cousin was not partnered with any terrorist group.* → *1*

Figure 3: Examples of usage data as training instances. Here affiliated is a CP and associated, merged, aligned, and partnered are the simpler options provided by 6, 2, 1, and 1 workers respectively.

More detailed statistics are shown in Table 1. From Table 1, we see that around 70% of the workers (mainly from India and the US) have successfully completed the task.

|           | #workers | #visitors |
|-----------|----------|-----------|
| instances | 18036    | 10758     |
| workers   | 164      | 71        |
| countries | 11       | 3         |

Table 1: Statistics of workers and simplification instances collected during all 9 iterations in the experiment. The column #workers shows the number of workers who have accepted and submitted the result while the column #visitors shows the number of workers who did not submit their results.

## 5 Learning-to-Rank

Learning-to-rank refers to a machine learning technique for training a model based on existing labels or user feedback for ranking task in areas like information retrieval, natural language processing, and data mining (Li, 2014). Learning-to-rank consists of a learning and ranking system. The system is trained by providing pairs of requests/queries and a target/ideal ranking for retrieved items. The learning model then constructs a ranking model on the basis of the training data ranking lists.

Ranklib[2], a well-known learning-to-rank library in Java from the Lemur Project is used to build the ranking models. Specifically, we have used the LambdaMART algorithm to train our learning and ranking models. LambdaMART combines LambdaRank and MART (Multiple Additive Regression Trees) (Burges, 2010; Donmez et al., 2009). While MART uses gradient boosted decision trees for prediction tasks, LambdaMART uses gradient boosted decision trees using a cost function based on NDCG for solving a ranking task.

---
[2]https://sourceforge.net/p/lemur/wiki/RankLib/

Normalized Discounted Cumulative Gain (NDCG) (Järvelin and Kekäläinen, 2002; Wang et al., 2013) is a family of ranking measures such as mean average precision (MAP) and Precision at K. NDCG is well-suited for our experiment for its capability of incorporating graded judgments. The graded judgments are obtained from the number of workers suggesting the candidate for the given CP target.

### 5.1 Features

To train the learning-to-rank model, we have designed a set of features that are important for text simplification. Based on the works of (Yimam et al., 2017; Horn et al., 2014), we have implemented the following list of features for the ranking model, which are partially derived from the candidate generating resources.

- **Frequency and length**: Due to the common use of these features in selecting the most simple lexical substitution candidate (Bott et al., 2012), we use three length features specifically the number of vowels, syllables, and characters and three frequency features: the frequency of the word in Simple Wikipedia, the frequency of the word in the document, and the frequency of the word in the Google Web 1T 5-Grams.

- **Lexical and distributional thesaurus resources**: We also use the number of similar words to the CPs and candidate suggestion based on lexical resources such as WordNet and distributional thesaurus as possible features. The features are normalized and scaled using the *featran's*[3] min-max scaler tool.

- **PPDB 2.0 and simple PPDB**: From the PPDB 2.0 and simple PPDB resources (cf. Sect. 4), we use associated scores as given by the resource, i.e.: *ppdb2score*, *ppdb1score*, *paraphraseScore*, and *simplificationScore*.

- **Word embeddings feature**: We use the Phrase2Vec embeddings as described in Section 4 to obtain vector representations for targets (CPs) and candidates. The cosine similarity of the candidates with the whole sentences as well as the cosine similarity of the candidates with the tri-gram words (one word to the left and one word to the right of the target CP) are used as features. The vector representations of the sentences and the tri-grams are the average of the individual vector representations of words in the sentences and the tri-grams.

## 6 Experiments

### 6.1 Baseline system

Our baseline system is built using a general purpose paraphrasing dataset from Yimam et al. (2016b). The dataset is based on essay sentences from the ANC[4] and BAWE corpora (Alsop and Nesi, 2009). We use the same feature extraction approach (see Section 5) for the development of the baseline model.

As it can be seen in Table 2, the results from each iteration are compared with the baseline system. We noted that the generic paraphrase datasets do not quite fit the task of text simplification as the need of the task is different. The lower performance on the baseline system can be attributed to the fact that the texts for the baseline system are collected from a different genre (essay sentences). We have to make clear that the first and all the subsequent iterations in the adaptive process do not use the baseline system.

### 6.2 Adaptive systems

We start with an empty ranking model (iteration 1), where candidates obtained from the resources are provided to the workers without an implied ranking. After collecting enough usage data from iteration 1, we trained a ranking model, which is used to re-rank candidates for the texts in the next iteration (iteration 2). Texts in iteration 2, which are exclusively different from those in iteration 1 are provided to the workers. Once workers completed the simplification task at iteration 2, we re-evaluate the ranking of candidates in iteration 2 based on the usage data (using NDCG@10 metric). An NDCG@10 score of 62.88 is obtained (see Table 2), which is already better than using the baseline system (60.66). Figure 4 shows the learning curve over the different iterations conducted in the experiment.

---

[3] https://github.com/spotify/featran
[4] http://www.anc.org/

| Testing | NDCG@10 | | | | | | | | | |
|---|---|---|---|---|---|---|---|---|---|---|
| | | | Training instances on previous iterations | | | | | | | |
| | #sentences | baseline | 1 | ≤ 2 | ≤3 | ≤4 | ≤5 | ≤6 | ≤7 | ≤8 |
| 1 | 115 | - | - | - | - | - | - | - | - | - |
| 2 | 214 | 60.66 | 62.88 | - | - | - | - | - | - | - |
| 3 | 207 | 61.05 | 63.39 | 65.52 | - | - | - | - | - | - |
| 4 | 210 | 58.21 | 60.73 | 65.93 | 67.46 | - | - | - | - | - |
| 5 | 233 | 56.10 | 62.53 | 65.66 | 66.00 | 70.72 | - | - | - | - |
| 6 | 215 | 62.18 | 61.05 | 66.51 | 67.86 | 69.88 | 72.36 | - | - | - |
| 7 | 213 | 57.00 | 62.07 | 64.02 | 64.88 | 67.28 | 69.27 | 74.14 | - | - |
| 8 | 195 | 56.56 | 59.53 | 62.11 | 63.03 | 64.54 | 67.40 | 71.05 | 75.83 | - |
| 9 | 224 | 56.14 | 63.48 | 65.58 | 65.87 | 69.18 | 69.51 | 71.31 | 71.40 | 75.70 |

Table 2: NDCG@10 results for each iteration of the testing instances using training instances from the previous iteration. For example, for testing at iteration 2, the NDCG@10 result using training data from the previous iteration, i.e. iteration 1, is 62.88. The baseline column shows the performance in each iteration using the generic paraphrasing dataset used to train the baseline ranking model.

| | Iteration 2 | | | Iteration 3 | | | Iteration 4 | | |
|---|---|---|---|---|---|---|---|---|---|
| Workers | Instances (#) | positive (%) | NDCG score | Instances (#) | positive (%) | NDCG score | Instances (#) | positive (%) | NDCG score |
| AXXXL5 | 950 | 10.21 | 51.35 | 2661 | 10.11 | 55.87 | 2771 | 9.78 | 56.57 |
| AXXX3N | 1591 | 10.31 | 45.45 | 3130 | 10.29 | 48.72 | 5367 | 10.23 | 47.98 |
| AXXXMY | 1117 | 10.12 | 55.15 | 2753 | 10.10 | 61.35 | 4809 | 10.13 | 63.76 |
| AXXXI7 | 70 | 10.00 | 49.33 | 2162 | 10.59 | 64.10 | 3988 | 10.38 | 66.82 |
| AXXX56 | 1190 | 10.42 | 54.63 | 2468 | 10.29 | 56.24 | 4477 | 10.27 | 58.79 |
| AXXXS1 | 824 | 10.19 | 54.45 | 1845 | 10.24 | 55.28 | 3045 | 10.15 | 58.78 |
| AXXXM9 | 448 | 10.04 | 55.25 | 896 | 10.04 | 56.00 | 2669 | 11.09 | 58.61 |
| AXXXAM | 1594 | 10.16 | 60.59 | 2999 | 10.17 | 61.96 | 4611 | 10.13 | 63.28 |
| AXXX3E | 615 | 10.73 | 59.44 | 1038 | 10.69 | 59.51 | 3451 | 10.66 | 62.59 |
| AXXXGI | 100 | 24.00 | 45.05 | 1979 | 11.22 | 56.72 | 3160 | 10.79 | 57.35 |

Table 3: The NDCG result for 10 different workers. *Instances* shows the total number of training instances used form the previous iteration while *positive* shows the total number of positive feedback provided by the user. The workers ID are obscured to protect their privacy.

Similarly, training instances collected from iteration 1 and iteration 2 are used to train a ranking model, which is used to re-rank candidates for texts in the next iteration (iteration 3). We continued the experiment for nine iterations and we record the performance at each iteration.

We have observed that the ranking model substantially improves on every iteration based on the NDCG@10 ranking evaluation measure. Table 2 also shows that if we test the performance on each of the models from the earlier iterations, the performance of the system declines, thus the system can make good use of more usage data if available. For example, on iteration 6, testing on a ranking model that is trained based on training instances from iteration 1 up to iteration 5 (≤**5**) produces an NDCG@10 score of 72.36 while testing on the ranking model trained based on training instances from iteration 1 up to iteration 4 (≤**4**) produces an NDCG@10 score of 69.88.

Furthermore, we have explored the effect of the adaptive system for individual workers. In this case, we have built simulated unique models for the 10 top workers, who have participated in at least 4 different batches (iterations) of the task. We use the first iteration the worker has participated as an initial model, and start using the model for the subsequent iterations. As we can see from Table 3 and Figure 5, the NDCG scores improve consistently over the iterations. The results also revealed that text simplification can be modeled differently based on the user needs (personalization).

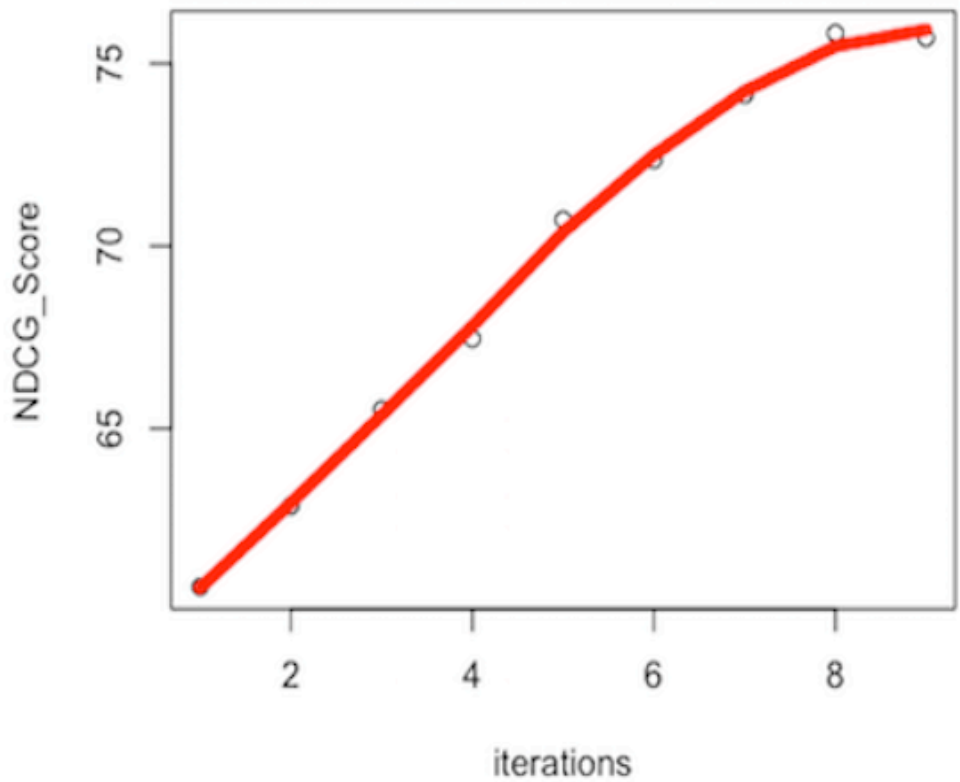

Figure 4: Learning curve showing the increase of NDCG@10 score over 9 iterations.

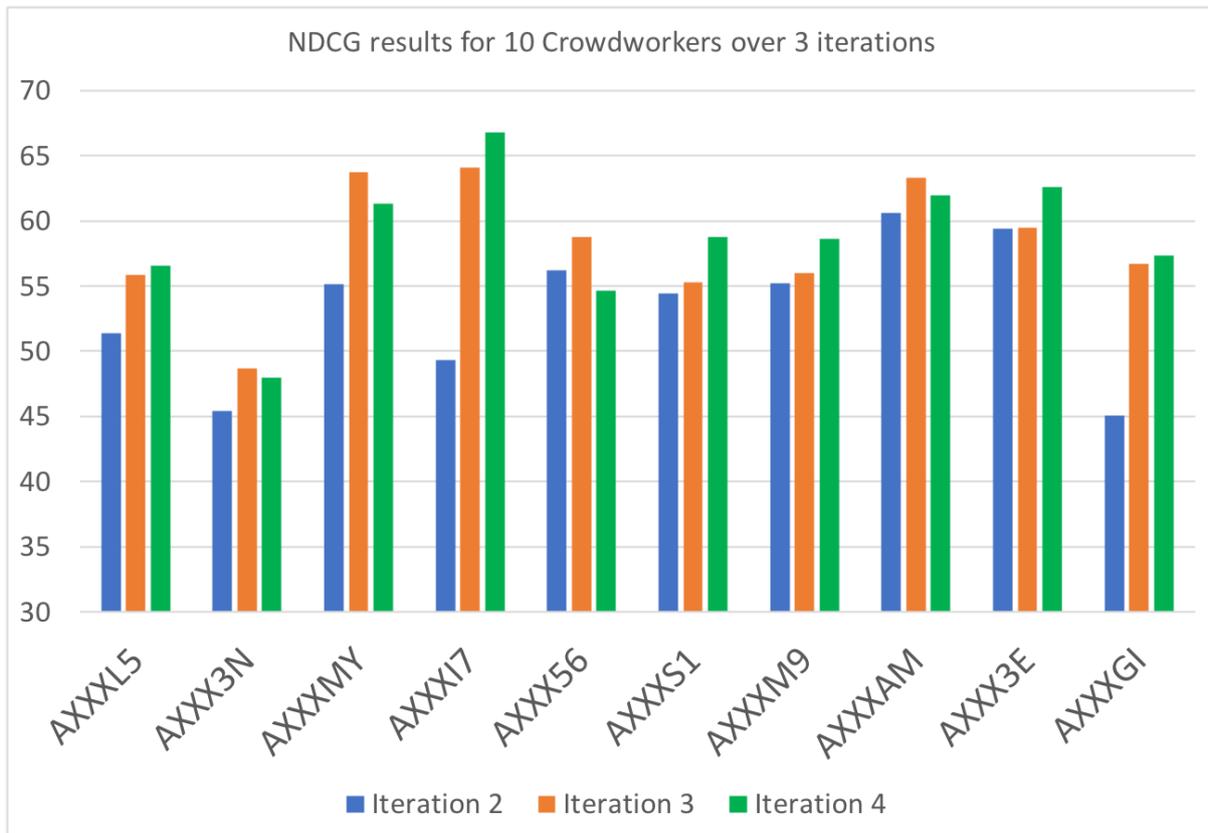

Figure 5: The increase of NDCG@10 score over 3 iterations for the top 10 workers ordered by their productiveness (who have have completed most HITs over several iterations).

## 7   Discussion

Most text simplification systems, and for that matter, most NLP models, are based on a traditional *collect and train* approach where first all the required training data are annotated , then training and evaluation is carried out after the data collection. To our knowledge, this experiment is the first scientific work to conduct an adaptive approach for text simplification where signals from usage data are collected in an interactive and iterative approach to improve the model of an NLP component.

We have demonstrated that our approach is noble in many aspects: 1) the adaptive learning model is integrated in the real-world NLP application (*live-usage – RQ1*), 2) the performance of the integrated adaptive model improves through usage data of the NLP application (*adaptability – RQ3*), 3) the integrated learning model potentially adapts to the needs of the user or user groups through usage data (*personalized NLP –RQ3*), and 4) we also have shown that adaptive systems can be evaluated incrementally, by comparing the system's suggestions by the ranking model to the actual ranking provided by the users (*incremental evaluation – RQ2*) .

In this research, we also have showcased how to perform web-scaled and real-time adaptive data collection using the Amazon MTurk crowdsourcing platform. The MTurk crowdsourcing platform has been mainly used to collect datasets for tasks that are not complex and difficult to complete such as identifying named entities or biomedical entities in a text, categorizing texts for spam, labeling an image with appropriate captions and so on. Using MTurk's external HIT, we are able to show that the MTurk crowdsourcing platform can be successfully used for complex NLP applications such as text simplification with a writing aid tool, which normally is limited to a lab-based experiment.

## 8   Conclusion and Future Directions

In this work, we have shown that the integration of an adaptive paraphrase ranking model effectively improves the performance of text simplification task. We have designed a full-fledged, web-scale based text simplification system where we have integrated an adaptive paraphrase ranking model into the tool.

Our tool is integrated with the Amazon Mechanical Turk crowdsourcing platform to collect usage data for text simplification.

To evaluate the performance of the adaptive system on the collected usage data, we have evaluated ranking model performance in an iterative way. In every iteration, we use the usage data exclusively from the previous iterations (except the first iteration that is used solely as a training data and we do not evaluate it) to training the learning-to-rank model. The result shows that, in every iteration, there is a large increase in performance based on the NDCG@10 evaluation metric.

We believe that this experiment is a showcase on how to develop a personalized NLP application. Using a similar approach, one can effectively deploy Par4Sim for a different purpose such as to write technical documents. The research also sheds light on a domain or task adaptions. One can use datasets collected for general purpose domains and it is possible to adapt the model based on the usage data over a period of time. This is a much cheaper alternative than collecting labeled datasets anew.

In the future, we would like to run a long-turn study with arbitrary users and arbitrary texts using a freely available online tool. Specifically for text simplification, the approach can be employed to provided graded complexity level of texts, as it is done for instance in the *Newsela* instructional content platform[5]. We also envision further possible tasks where adaptive learning helps, such as collaborative text composing and recommender systems.

The software is openly available under ASL 2.0 license and the resources and datasets used in this paper are released under CC-BY[6]. The demo of the tool as it was used inside the MTurk browser can be accessed online.[7]

---

[5] https://newsela.com/data/
[6] https://uhh-lt.github.io/par4sim/
[7] https://ltmaggie.informatik.uni-hamburg.de/par4sim/

## Acknowledgement

This work has been partially supported by the SEMSCH project at the University of Hamburg, funded by the German Research Foundation (DFG).

We would like to thank the PC chairs, ACs, and reviewers for their detailed comments and suggestions for our paper. We also would like to thank colleagues at LT lab for testing the user interface. Special thanks goes to Rawda Assefa and Sisay Adugna for the proofreading of the Amharic abstract translation.